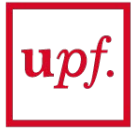 Treball de fi de màster de Recerca

# Text-To-Speech Data Augmentation for Low Resource Speech Recognition

**Nom i Cognoms** Rodolfo Joel Zevallos Salazar

**Màster:** Lingüística Teòrica i Aplicada

**Edició:** 2020-2021

**Directors:** Dra. Núria Bel

**Any de defensa:** 2021

**Col·lecció: Treballs de fi de màster**

**Departament de Traducció i Ciències del Llenguatge**

# Acknowledgements

I would especially like to thank Núria Bel for guiding me throughout this project, and helping me to improve every day as a researcher. I would also like to thank Mireia Farrús, Guillermo Cambara and Alex Peiro for their constant support in the development of the speech technologies performed in this project. I would also like to thank my family for always being with me and supporting me to achieve my goals. Finally, I cannot forget to mention Eva for her energetic, philosophical and emotional help throughout this project.

# Abstract

Nowadays, the main problem of deep learning techniques used in the development of automatic speech recognition (ASR) models is the lack of transcribed data. The goal of this research is to propose a new data augmentation method to improve ASR models for agglutinative and low-resource languages. This novel data augmentation method generates both synthetic text and synthetic audio. Some experiments were conducted using the corpus of the Quechua language, which is an agglutinative and low-resource language. In this study, a sequence-to-sequence (seq2seq) model was applied to generate synthetic text, in addition to generating synthetic speech using a text-to-speech (TTS) model for Quechua. The results show that the new data augmentation method works well to improve the ASR model for Quechua. In this research, an 8.73% improvement in the word-error-rate (WER) of the ASR model is obtained using a combination of synthetic text and synthetic speech.

# Table of Contents

# 1. Introduction

At present, half of the languages spoken in the world are endangered (Austin & Sallabank, 2011). The use of communication and information technologies (Adda et al., 2016; Stuker et al., 2016) plays an important role in the preservation of these languages (Bird, 2020; Blachon et al., 2016).

The future of informatics is intertwined with language technology. New artificial intelligence and deep learning technologies make it possible to use vast collections of informational, textual, and other linguistic data to create linguistic tools that are widely used in many domains, such as business, finance, social networks, or security systems. Automatic dialog systems can improve services in many areas of daily life. For example, virtual assistants such as Microsoft's Cortana, Amazon's Alexa or Apple's Siri have become an important part of people's daily lives.

As part of the dialog systems there are ASR systems, which convert speech signals into writing, and TTS systems, which allow to generate speech from text. Both systems are of vital use for endangered languages, as they do not only give the language an advantage, but can be crucial for its survival (Camacho & Zevallos, 2020).

In recent years, deep learning techniques have been applied to ASR and TTS systems with ever increasing computational power. Thus, some encouraging results have been obtained in these systems (Gulati et al., 2020, Łańcucki et al., 2020). However, the best performing ASR[1] models for English required no less than 1,000 hours of fully transcribed audio (Synnaeve et al., 2019; Gulati et al., 2020). Besides, the Deep Speech 2 model (Amodei et al., 2016) used up to 10,000 hours[2] of transcribed audio for getting best results. Therefore, current state of the art for ASR models, increases the importance of using corpora with large amounts of fully transcribed audios for developing competitive ASR models. On the other hand, languages that do not count with a sufficient number of transcribed audio to obtain good results using these models have two options: to obtain more transcribed audios by recording native speakers or using massive corpus collection platforms, which significant amount of work and money, or explore other alternatives that do not require a large investment of money and time, to obtain a greater amount of transcribed audio.

---
[1] https://github.com/syhw/wer_are_we

In this research, a new data augmentation method is proposed to improve ASR models for agglutinative and low-resource languages. We propose that this data augmentation method be for agglutinative languages in order to contribute to the research being carried out in favor of native and indigenous languages of the American continent (Mager et al., 2018), most of which are agglutinative and low-resource languages. For this research, the Quechua language will be used, since it is agglutinative and is considered a low-resource language.

## 1.1. Goals

The creation of new data augmentation methods for low-resource languages is becoming increasingly investigated. Automatic speech recognition based on deep learning require large amounts of transcribed data for creating models. For low-resource languages, models created with a limited amount of data are either inadequate or their performance lags far behind that of languages with large amounts of data. However, a more pragmatic and fruitful approach is to explore different ways to generate new transcribed data synthetically.

---
[2] https://www.openslr.org/12

**General Objective:**

To demonstrate that our new data augmentation method based on a seq2seq model to generate synthetic text and a TTS model to generate synthetic audio improves the performance of the ASR models of Quechua.

**Specific Objectives:**

- Increase the quantity and improve the quality of the speech corpus for Southern Quechua.
- Develop a seq2seq data augmentation model for Southern Quechua that generates synthesized text.
- Develop a TTS model for Southern Quechua that generates that generates synthesized audio.
- Develop an ASR model for Southern Quechua

## 2. Quechua language

This section presents a brief introduction to the main characteristics of the language that was selected to perform the experiments of the new data augmentation method to improve ASR models for agglutinative languages with low resources.

Quechua is a family of languages spoken in South America with about 10 million speakers, not only in the Andean regions but also in the valleys and plains connecting the Amazon jungle and the Pacific coast. Quechua languages are considered highly agglutinative with a subject-object-verb (SOV) sentence structure as well as mostly postpositional.

Although the classification of Quechua languages is still open to research (Heggarty et al., 2005; Landerman, 1991), works on language technology for Quechua (Ríos, 2015; Ríos and Mamani, 2014) have adopted the categorization system described by Torero (1964). This categorization divides the Quechua languages into two main branches, QI and QII (see Figure 1). The QI branch corresponds to the dialects spoken in central Peru. The QII branch is further divided into three branches, QIIA, QIIB and QIIC. QIIA groups the dialects spoken in northern Peru, while QIIB groups those spoken in Ecuador and Colombia. In this paper, we focus on the QIIC dialects, which correspond to those spoken in southern Peru, Bolivia and Argentina. Mutual intelligibility between speakers of QI and QII dialects does not always occur. However, the QII dialects are close enough to each other to allow mutual intelligibility.

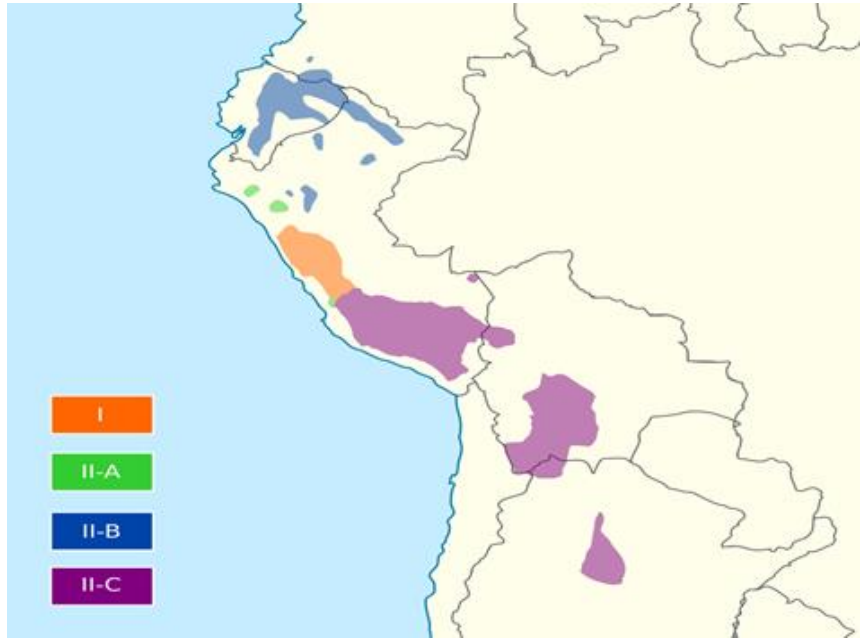

**Figure 1** The four branches of Quechua language

## 2.1. Orthographic Variation

In order to address orthographic variation, experts proposed several writing standards, the most notable being that of Cerrón-Palomino (1994), currently adopted as the Quechua writing standard by the Ministries of Education of Peru and Bolivia (with a minor modification with respect to that of Peru). However, there are still some discrepancies with the NGO Academia Mayor de la Lengua Quechua in Cusco, which dares to disseminate its own orthography.

Table 1 shows the dialectal variations of suffixes and their corresponding standard form (Rios, 2014).

**Table 1** Suffix variation and normalization.

| Function | Variations | Standard |
|---|---|---|
| Progressive | -chka / -sha/ -sa/ -sva | -chka |
| Genitive (after a vowel) | -p / -q / -h / -j | -p |
| Evidential (after a vowel) | -m / -n | -m |
| Additive | -pis/ -pas | -pas |
| Euphonic | -ni / -ñi | -ni |
| Plural forms (1st and 2nd person) | -chis / -chik / -chiq | -chik |
| Assistive | -ysi / -schi / -scha | -ysi |
| Potential forms | -swan / -chwan | -chwan |

## 2.2. Suffix classes

At present, 130 suffixes are recognized as common to all dialects, although the exact number and morphosyntactic behavior vary between dialects and regions. The same suffix may have different forms between dialects, but still fulfill the same function. In regions where the language had (and still has) interaction with other native languages, such as Aymara, the adoption of suffixes and vocabulary from these languages over the years was inevitable. Today, 5 functional classes of suffixes can be distinguished:

- Nominalizing: they transform a verb into a substantive verb (verb -> subst)
- Verbalizing: they transform a substantive into a verb (subst -> verb)
- Nominal: they modify a substantive (subst -> subst)
- Verbal: they modify a verb (verb -> verb)
- Independent: they can modify a verb or a substantive.

## 2.3. Phonetics and phonology

Chanca Quechua has a total of 15 consonants, most of them voiceless, as shown in Table 2. As in Spanish, the phoneme [tʃ] is written as ch, [ɲ] as ñ, and [ʎ] as ll. Collao Quechua also has a glottal and an aspirated version of each plosive consonant, leading to a total of 25 consonants. However, the use of voiced consonants present in the phonemic inventory of Spanish is common in all dialects due to the large number of borrowings present.

In both dialects, only [æ], i [ɪ] and u [ʊ] are encountered as phonemic vowels, although in the proximity of [q], its allophones [ɑ], [ɛ] and [ɔ], respectively, are pronounced. The OMNIGLOT online database has entries for standard chanca[3] and collao[4].

**Table 2** Consonants in the phonemic inventory of quechua chanca (IPA)

|  | **Bil** | **Alv** | **Pal** | **Vel** | **Uvu** | **Glo** |
|---|---|---|---|---|---|---|
| Plosive | p | t | tʃ | k | q |  |
| Nasal | m | n | ɲ |  |  |  |
| Fricative |  | s |  |  |  | h |
| Lat. Approx. |  | l | ʎ |  |  |  |
| Approximant |  | ɹ |  |  |  |  |
| Semi-consonant | w |  | y |  |  |  |

## 2.4. Language Technologies and Resources for Quechua

Regarding technologies developed with text resources, the Instituto de Lengua y Literatura Andina Amazónica (ILLA) worked on the construction of electronic dictionaries for Quechua, Aymara and Guaraní, which are currently used as a parallel corpus for the creation of automatic Spanish-Quechua translators. In 2008, the first parallel treebank for Quechua-Spanish was developed (Rios, 2008). The Hinantin group of the Universidad Nacional San Antonio Abad del Cusco (UNSAAC - Peru) produced a spell checker for LibreOffice (Rios et al., 2011). Rios (2016) developed a hybrid machine translator for Spanish-Quechua, a chunker and a morphological analyzer for Southern Quechua. The AVENUE-Mapudungun project developed a machine translation system for Quechua and Mapudungun (Monson et al., 2006); Ortega et al. (2020) developed a machine translator for Quechua with several morphological segmentation techniques and a new one for decomposing suffix-based morphemes.

---

[3] https://www.omniglot.com/writing/ayacuchoquechua.htm
[4] https://www.omniglot.com/writing/quechua.htm

Regarding linguistic technologies developed with speech resources, in 2012 the Hinantin group of the Universidad Nacional San Antonio Abad del Cusco (UNSAAC - Peru) developed a TTS model of Southern Quechua with 15 hours of transcribed audio that they collected (Vargas et al., 2012). The TTS model was built using the festival[5] tool. In 2018 the Siminchikkunarayku project developed a corpus of 97.5 hours of fully transcribed audio for Southern Quechua (Cárdenas et al., 2018), based on crowdsourcing methodology, which was used for the construction of a 5-gram language model using the K-Nearest Neighbor (KNN) algorithm and an ASR model using the HTK[6] tool. That same year, an ASR model exclusively for numbers was developed using Mel-Frequency Cepstral Coefficients (MFCC), Dynamic Time Warping (DTW) and KNN (Chacca et al., 2018) achieving a WER of 8.9% with a corpus of 5 hours of transcribed audio. On the other hand, in 2019, an ASR model was developed using the HTK tool, based on the monophone HMM topology, which is a model with five states per HMM and no jumps that is considered the best in terms of accuracy (Zevallos et al., 2019). This model was trained using the corpus presented by Cárdenas et al. (2018) achieving a WER of 12.7%.

---

[5] https://web.archive.org/web/20100925093527/http://www.cstr.ed.ac.uk/projects/festival/
[6] https://htk.eng.cam.ac.uk

Finally, that same year, Aimituma et al. (2019) presented the first ASR model based on deep learning and developed with the Kaldi[7] tool, and trained with a corpus of 18 hours of transcribed audio created by themselves, achieving a WER of 40.8%.

On the other hand, as part of this thesis and the "Lurin-corpus" project of the Pontifical Catholic University of Peru (PUCP), a new corpus for Southern Quechua was elaborated. For this new corpus, intensive linguistic work was carried out to build a total of 8,000 phrases. The phrases were collected from the Qullaw-Spanish and Chanka-Spanish dictionaries published by the Peruvian Ministry of Education (2021), from Soto's Quechua-English functional dictionary (2007) and from the book Autobiografía de Condori (2013). These phrases as a whole include all the phonemes of the Southern Quechua language (Chanca and Collao); in addition, they were used to create base audios for the massive corpus collection application called "Huqariq"[8] (Camacho et al., 2020). They were then standardized to Southern Quechua as was done by Cárdenas et al. (2018) using the Ríos (2016) normalizer. They were recorded by native speaker linguists of each dialectical variety of Quechua (Chanca and Collao) using "Tarpuriq", an application that was developed exclusively for this task. Each recorded sentence lasted approximately five seconds, for a total of about 11 hours for the entire dataset. The recordings made by the linguists were used as prompts in the Huqariq application. Huqariq users had to repeat the audios they listened to. Users were not exposed to all 8,000 sentences, but only to a random sample of 240 (1,200 seconds, 20 minutes). Each user took approximately one and a half hours to complete the task. The recordings made by users in Huqariq were preprocessed by the quality enhancement algorithms that the application features, which eliminates excess background noise, excess non-speech audio, audios that feature background music, and audios that are not spoken in Quechua, the latter feature being added thanks to the first automatic speech recognition developed by Zevallos et al. (2019). Finally, the recordings made by Huqariq users and those made by linguists in Tarpuriq are automatically aligned with their respective phrases. The corpus developed in this research reached 83.3 hours of fully transcribed audio. The statistics of the speech corpus developed for this research are shown in the following table.

---

[7] https://kaldi-asr.org
[8] https://play.google.com/store/apps/details?id=com.itsigned.huqariq&hl=es_419&gl=US

**Table 3** Speech Corpus Statistics of the Lurin-corpus

|  | **Female** | | **Male** | |
|---|---|---|---|---|
|  | Quechua Chanca | Quechua Collao | Quechua Chanca | Quechua Collao |
| #Speakers | 24 | 72 | 16 | 48 |
| #Length | 21.3h | 45.2h | 7.1h | 9.7h |

The new corpus developed in this thesis, and the corpus and language model presented by Cardenas et al. (2018) will be used in this research. These resources will be used to develop the ASR and TTS models as part of the new data augmentation method for agglutinative and low-resource languages.

# 3. State of The Art

This section describes the models and methods used at present in developing data augmentation tools, ASR and TTS based on deep learning, as well as the methods for accuracy evaluation of these models. Finally, the available resources for Quechua are described.

## 3.1. Automatic Speech Recognition (ASR)

In the field of automatic speech recognition, modular and neural approaches are used to develop these technologies. These approaches generally require 1,000 (Panayotov et al., 2015) hours of transcribed audios to obtain excellent results. These transcribed audios are recordings made by people of the same language and aligned with their respective transcription. These approaches employ some evaluation method to verify the efficiency of the models. The evaluation method used by most research is the so-called word-error-rate (WER), which quantifies the proportion of words that would have to be replaced, deleted or inserted in the transcription made by the model to match the original transcription. However, there are other evaluation methods such as token-error-rate (TER), which quantifies in the same way as WER, but at the token level. TER is more suitable for agglutinative languages, while WER is more suitable for inflectional languages.

Models based on the modular approach frequently use the Switchboard corpus (Godfrey et al., 1992), which contains 260 hours of transcribed audio for English. The main models that stand out in this approach are those of IBM (Chen et al., 2006), BBN (Matsoukas et al., 2006) and Microsoft (Seide et al., 2011), which obtained a WER of 13%, 13.5% and 16.1% respectively.

In the last decade, models based on artificial neural networks (ANNs) presented a powerful replacement to the modular approach (Maas et al., 2014). An important breakthrough is the Deep Speech 2 model published by Baidu (Amodei et al., 2016), which implements a complex architecture using convolutional networks, recurrent networks and fully connected networks, and obtained a WER of 5.82% training with a total of 10,000 hours of transcribed English audio. On the other hand, a significant improvement in neural network-based ASR was obtained by the Wav2letter++ system, published by Facebook (Pratap et al., 2018) which obtained a WER of 4.81% using the Wall Street Journal (WSJ) speech corpus[9] (Paul et al., 1992) with a total of 80 hours of transcribed speech. The Wav2letter++ model implements a novel optimization feature that does not require the audio to be aligned with the text at the word level, just like the well-known connectionist temporal classification (CTC) (Graves et al., 2006). This model is considered an excellent choice for developing ASR models for low-resource languages.

In this research, the Wav2letter++ model (Pratap et al., 2018) will be used to build an ASR model for Quechua as part of the experiments.

## 3.2. Data Augmentation

Data augmentation methods are techniques that generate synthetic data from an existing dataset and are used to generate both text and audio data to cope with the problems that deep learning methods face when creating models with limited training datasets. The lack of corpora with a large number of hours of transcribed audio for languages other than English has sparked the creation of methods to solve this problem.

---

[9] https://catalog.ldc.upenn.edu/LDC93S6A

### 3.2.1. Text

Data augmentation techniques to generate synthetic text have been gradually explored to improve machine translation systems, question-answering, dialog systems (chatbot, agent-based) among others. The lack of labeled monolingual text, or parallel text, has led researchers to seek new alternatives to improve text-based natural processing models. In 2018, a study generated synthetic texts from French to English translations using a machine translation model (Yu et al., 2018). Kobayashi (2018) developed a method that replaces words with synonyms to generate synthetic text. On the other hand, Hou et al. (2018) presented a data augmentation methodology based on the seq2seq model to generate synthetic text by replacing words with words from the same semantic field. This methodology incorporates a delexicalization method, which replaces segments of a sentence with labels based on their semantic frame (Xie et al., 2013); for example, the delexicalization method transforms the following sentence "Show me the nearest restaurant" by replacing the word "restaurant", with a <place> labelslot, while the word, "nearest", with a <distance> label In addition, this methodology incorporates a diversity classification method, which verifies that when using the seq2seq model to generate new delexicalized sentences, they are not too similar to the original sentence. This method helps to generate greater sentence diversity while maintaining the same semantic field of the original sentences. For example, the diversity ranking method works as follows: if three delexicalized sentences, "is there a <place><distance>?", "tell me the <place> most <distance>" and "find me the <place> <distance>" are generated by the seq2seq mode, the method ranks them based on the least similar to the original sentence. Finally, as the last step of the methodology of Hou et al., the seq2seq model is applied to generate new delexicalized sentences using the delexicalized sentence obtained in the delexicalization step. In addition, the label of the new delexicalized sentences are replaced with words from the same semantic field. This methodology was used to improve the accuracy of language comprehension models in a task-oriented dialog system. The methodology of Hou et al. improved the system by 6.38%. In 2019, a set of universal data augmentation techniques called Easy Data Augmentation (EDA) was developed by Wei et al. (2019). This data augmentation method employs four operations (synonym substitution, random insertion, random swapping, and random deletion) to create synthetic text, improving text classification models by 0.8% accuracy.

In this research, as the basis of our new data augmentation method, we will use the steps of the methodology of Hou et al. (2018), which was described earlier in this section, modifying the delexicalization method so that it can be used efficiently for agglutinative languages, since the delexicalization method proposed by Hout et al. uses a word-level tokenization method, part-of-speech tagging, and lemmatizer for inflectional languages, in addition to not incorporating a solution in the tagging of semantic frames of words not found in the lexicon.

### 3.2.2. Audio

ASR models require large amounts of data (Amodei et al., 2016) and currently can achieve good results for high-resource languages (English, Mandarin, and Spanish). Data augmentation methods for ASR modeling are receiving more and more attention from researchers. Among the speech data augmentation techniques are those that augment the data by audio distortions; that is, by modifying the audio speed, adding background noise or distorting the speech, those that generate new audio using a TTS model, and those that modify the architecture of the ASR model.

Audio distortion methods have been evolving to improve data quality. In 2014, a data augmentation technique for low-resource languages was presented by distorting the vocal tract length (VTLP) in the data (Ragni et al., 2015) managing to improve ASR models by 2.5% in token-error-rate (TER). Furthermore, Cui et al. (2015) presented a method mixing vocal tract length distortion and stochastic feature mapping (SFM), managing to improve ASR models by 3.2% in WER. In 2020, a substantial improvement in this technique was achieved by distorting noise addition, velocity adjustment and pitch shifting in the original audios (Lu et al., 2020) managing to reduce the WER by 5.1%. Currently, data augmentation techniques using a TTS model are giving good results, Rygaard et al. (2015) used a TTS model to generate new audios and improve their ASR for low-resource languages, achieving a 20% reduction in WER. In 2019, a technique combining audio distortion and data augmentation technique using Google translate TTS (gTTS) was developed for Turkish (Gokay et al., 2019) obtaining a 14.8% reduction in WER. Finally, data augmentation techniques by modifying the ASR model have more and more relevance. Park et al. (2019) presented SpecAugment, a data augmentation method that modifies the spectrogram image of the original audio by masking and image warping techniques, achieving a 6% reduction in WER. In 2020, inspired by the SpecAugment model, new data augmentation methods were developed based on this model (Nguyen et al., 2020).

In our research, as part of our new data augmentation method, data augmentation techniques using a TTS model for Quechua will be used to generate synthetic speech. On the other hand, the data distortion methods proposed by Lu et al (2020) will be used to compare them with the new data augmentation method.

## 3.3. Text-to-Speech (TTS)

Over time in the development of text-to-speech, different techniques have been developed. Concatenated synthesis with unit selection (Hunt et al., 1996; Black et al., 1997) was the most widely used technique for years until 2005. Then, the statistical parametric technique was developed (Tokuda et al., 2000; Zen et al., 2009), which directly generates smooth trajectories of speech features to be synthesized by a vocoder. In 2016, WaveNet (Van den Oord et al., 2016) a generative model of waveforms in the time domain was developed, which produces very human-like audio quality. Building on new models based on deep learning, in 2017, Tacotron (Wang et al., 2017), a seq2seq architecture for producing magnitude spectrograms from a sequence of characters, was developed. It simplifies the traditional speech synthesis process by replacing the production of these linguistic and acoustic features with a single neural network trained solely from the training audio. To vocalize the resulting magnitude spectrograms, Tacotron uses the Griffin-Lim algorithm (Tamamori et al., 2017) for phase estimation, followed by a short-time inverse Fourier transform. As the authors point out, this was simply a placeholder for future neural vocoder approaches, as Griffin-Lim produces characteristic artifacts and inferior audio quality compared to approaches such as WaveNet. Finally, Shen et al. (2018) presented Tacotron 2, an end-to-end model that combines the Tacotron model with a modified WaveNet vocoder (Sotelo et al., 2017), achieving a very natural voice with hours transcribed audio.

In this research, as part of the new data augmentation technique, the Tacotron 2 model (Shen et al., 2018) will be used to create the TTS model for Quechua as it generates a more natural voice and can be used to generate new audios with good quality that can be used for enlarging the ASR training corpus.

# 4. Methodology

## 4.1. Data augmentation method

This section details the new data augmentation method for improving ASR models for agglutinative and low-resource languages. Our new data augmentation method aims to create synthetic text from a seq2seq model using a delexicalization method for agglutinative languages, which will be developed in this section. In addition, these new synthetic texts will be used as inputs in a TTS model to generate synthetic audio. Both synthetic text and synthetic audio will be aligned to train the ASR model for agglutinative languages.

Importantly, our new data augmentation method is novel because, unlike other data augmentation methods for improving ASR models (Lu et al., 2020; Park et al. 2019; Rygaard et al. 2015), it combines for the first time a data augmentation method for text and a data augmentation method for speech to improve the results of ASR models for agglutinative and low-resource languages.

The proposed method consists of two main steps:

1) Synthetic text generation: Synthetic text generation: the synthetic text is created using a modified version of the methodology proposed by Hou et al. This modified version incorporates a new delexicalization method for agglutinative languages developed in this research.

2) Synthetic audio generation: the synthetic text obtained in step 1 is used as input to generate synthetic audio using a TTS model (Shen et al., 2018).

This novel data augmentation method proposes a new system for agglutinative languages based on the delexicalization method proposed by Hou el at. 2018. This new delexicalization method is used to generate delexicalized sentences which in turn will be used in a seq2seq model to generate synthetic texts. The synthetic texts generated from the seq2seq model are used to generate synthetic audios using a TTS model. Another important use of synthetic texts is to improve the language model to be used for the ASR model. Our data augmentation method is sequential, i.e., it is required to use the text data augmentation method with our new delexicalization method to use the speech data augmentation method then.

The general process of our new data augmentation method is shown below.

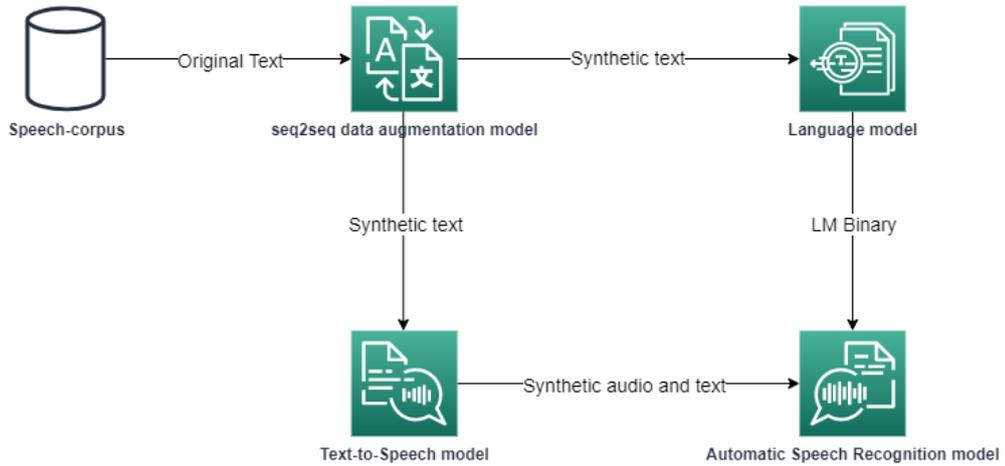

**Figure 2** General process of the new data augmentation method and how it is used for generating the language model and the ASR model.

The following two subsections describe these two steps in detail. As will be shown in Section 4.3, the performance of ASR models for low-resource agglutinative languages trained with synthesized data increases positively.

### 4.1.1. Synthetic text generation

Synthetic text generation for agglutinative languages is achieved using the methods of delexicalization, diversity classification, text generation using a seq2seq model and data surface realization proposed by Hou et al. (2018) but changing the delexicalization method to be used for agglutinative languages.

Figure 3 shows the steps to be followed to achieve synthetic text generation.

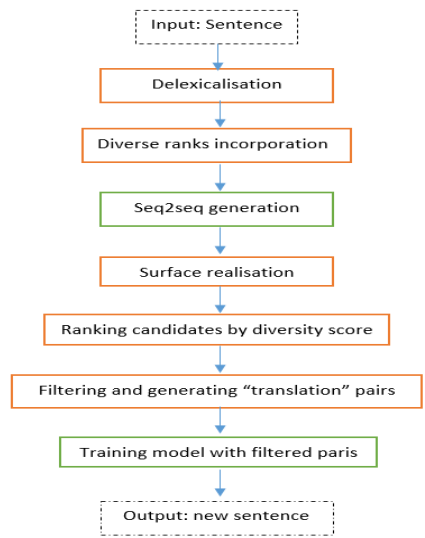

**Figure 3** The steps to generate synthetic text. (Hou et. al, 2018)

**New delexicalization method**

We create a new delexicalization system for agglutinative languages different from the one proposed by Hou et al. (2018). This new method is developed using Quechua resources.

To create delexicalized sentences, a lexical database is needed to provide the semantic frames of words in labels such as <place> or <time>. For this case, we use the Quechua lexicon proposed by Rudnick (2011), which provides the semantic frames of words in Southern Quechua. In order to tag the sentences of the training dataset with their semantic frames, we normalize the sentences using the tokenizer, POS Tagging and the lemmatizer of Quechua developed by Ríos (2015); this step of normalizing the sentences is different from the one performed by Hou et al. since they use a tokenizer, a POS Tagging and a lemmatizer for inflectional languages. Here is an example of lemmatizing and POS Tagging of a word in Quechua. The Quechua word "wasichapi", meaning "in the little house", is lemmatized to its base word "wasi" meaning "house" and tagged with "NOUN" while the suffixes "cha" (diminutive) and "pi" (locative) are tagged with "SUF-DI" and "SUF-LO" respectively.

After normalizing all sentences in our dataset, we label the words using Rudnick's (2012) lexicon. Our method incorporates a solution for low-resource language lexicons wherein case a lemmatized word is not found in the lexicon of use; bilingual dictionaries can be used to look up its corresponding to the lemmatized word. In our case, we used the Quechua-English bilingual dictionary of Soto (2007) to look up some lemmatized words that were not found in Rudnick's lexicon. Once those words were found, we used the English lexicon provided by the NLTK[10] Python library to tag those word and put those tags in our Quechua lemmatized words.

Our delexicalization method selects the labels from the most frequent semantic frames in all sentences. This step is important since we only consider the labels that appear most frequently in all sentences, as it is necessary to obtain as many delexicalized sentences with the same semantic frame label as possible. In our case, our method found that most sentences had the semantic frame labels of month_name, city_name and time_name. Below is an example of the labeled words with their respective semantic frames.

```
qanyanwata    B-date month_name
riqira    0
juliacachu    B-fromloc city_name
achka    0
wambrakunata    0
kaywata  B-date month_name
pitasi    0
riqiraqchu    0
kay 0
sullanachu    B-toloc city_name
```

**Figure 4** Example of delexicalization of a Quechua utterance, where the words on the right are the labels of the slots and the words on the left are the labeled words.

---

[10] http://www.nltk.org

Finally, the words that have the semantic frame tags selected by our method are replaced by slots with the same semantic frame tag name and in turn, these words are added to a vocabulary by semantic frame tag type, which will be used in the process of filling the delexicalized sentences generated by the seq2seq model.

The following is an example of a delexicalized sentence: "<B-date month_name> riqira <B-fromloc city_name> achka wambrakunata <B-date month_name> pitasi riqiraqchu kay <B-toloc city_name>"

### 4.1.2. Synthetic audio generation

In this step for synthetic audio generation, we train a TTS model for Quechua based on the Tacotron 2 model (Shen et al., 2018) in order to achieve good performance with 15 hours of transcribed audio recorded by a single speaker from the corpus used in this research and reaching a minimum of 3.2 in MOS (understandable audio). Subsequently, we used the synthetic texts generated in the first step of our data augmentation method to generate synthetic speech from our TTS model of Quechua.

## 4.2. Corpus

One of the most important requirements for the development of ASR models is undoubtedly the set of transcribed audios, which are made from sentences extracted from different written texts. Moreover, the audios have to be phonetically rich and balanced, phonetically rich meaning that all the phonemes of the language are present, and balanced meaning that the phonemes should, as far as possible, have the same number of occurrences. This section describes the corpora used for this research.

The speech corpus used for this research is the developed by Cárdenas et al. (2018) for Southern Quechua (Chanca and Collao), which has a total of 97.5 hours of fully transcribed audio, and Lurin corpus developed in this research for Sourthern Quechua (Chanca and Collao), which has a total of 83.3 hours of fully transcribed audio.

In order to improve the quality of the audios of the two aforementioned corpus, a pre-processing was carried out, where the audios containing music and parts only in Spanish were eliminated. In addition, the audios longer than 30 seconds were divided into segments no longer than 30 seconds. Finally, the audios were transformed to mono channel, 16 kHz sampling, 16-bit precision encoding and WAV format.

Below are the statistics of the corpus to be used for the experiments in this research, which is a combination of the corpus by Cárdenas et al (2018) and Lurin corpus developed in this research.

**Tabla 4** Speech corpus statistics of experiment

|        | **Total** | **Training** | **Validation** | **Test** |
|--------|-----------|--------------|----------------|----------|
| Length | 123.75h   | 99h          | 12.4h          | 12.35h   |

## 4.3. The experiment

In order to test our new data augmentation method that was intended to improve the ASR models for low-resource agglutinative languages, the resources of the Quechua language were used, since it is a low-resource agglutinative language. Fourth experiments were conducted, in order to compare our proposed new method with a baseline and other data augmentation methods.

### 4.3.1. First experiment

In the first experiment one ASR model was trained for Quechua as baseline. The ASR model was trained with the training data shown in the table 4. It has a total of 99 hours of transcribed audio from Quechua.

The ASR model of Quechua is based on the Wav2letter++ model (Pratap et al., 2018). The model was implemented from Facebook's wav2letter++ repository[11], which is developed entirely in Torch7[12] and in C programming language.

The model was trained for 242 epochs. Stochastic gradient descent with Nesterov momentum was used along with a minibatch of 4 expressions. The learning rate was 0.002 to achieve faster convergence and it was annealed with a constant factor of 1.2 after each epoch, in addition a momentum of 0 was used.

---

[11] https://github.com/flashlight/wav2letter
[12] http://torch.ch

The architecture of the Quechua ASR model consists of 2 convolutional layers with stride, 12 1D convolutional and 2 convolutional layers with kw = 1, which are equivalent to fully connected layers (see Figure 5). All layers have a dropout of 0.25, except the first one. The ASR system was trained using CTC which is the core of the Deep Speech architecture (Amodei et al., 2016).

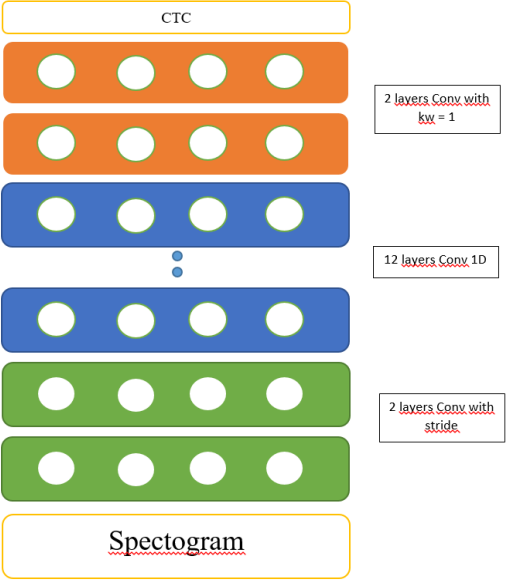

**Figure 5** Wav2letter architecture used in our approach

A vocabulary was created containing 34 graphemes: the standard Spanish alphabet plus the apostrophe, silence and 6 graphemes of vowels with accents and /ü/, plus a lexicon with all the words of the corpus separated at letter level.

The hyperparameters of the architecture, as well as those of the decoder, were adjusted using the validation set. In what follows, we report token error rates (TER) or word error rates (WER). The WERs were obtained using the decoder presented by Collobert et al. (2016).

The MFCC features were calculated with 13 coefficients, a 25 ms sliding window and a 10 ms interval. First and second order derivates were included. Power spectrum features were calculated with a 25 ms window, 10 ms interval and 257 components. All features are normalized (mean 0, std 1) per input sequence.

The baseline ASR model of Quechua in this first experiment achieved a WER of 31.48% with a total of 99 hours of transcribed audio.

### 4.3.2. Second experiment

In the second experiment, the method of data augmentation through audio distortion proposed by Lu et al (2020) was used to augment the speech data of the Quechua corpus. The nlpaug[13] library was used to manipulate the training audios (99 hours) by modifying the speed according to a randomly selected coefficient in the range between 0.85 and 1.15, which is where this data augmentation technique performs best.

By applying this speed distortion to the training audios, we were able to duplicate the training data to 198 hours of transcribed audio.

The ASR model of Quechua was re-trained with the same configuration as the base model, obtaining a WER of 25.13.

---

[13] https://github.com/makcedward/nlpaug

### 4.3.3. Third experiment

In the third experiment, our new data augmentation method was tested. As the first step of the method, the synthetic text augmentation repository[14] developed in this research was used. The seq2seq model was trained with the delexicalized training audio transcripts. The total number of delexicalized transcripts was 8,320 sentences. OpenNMT (Klein et al., 2017) was used as the implementation of the seq2seq model to generate synthetic text. The number of layers in LSTM was set to 2 and the size of hidden states to 1000. The same steps as Luong et al. (2015) and Kingma et al. (2014) were followed to train the seq2seq model. The dropout of the seq2seq model was set to {0, 0.1, and 0.2} considering its regularization power on small data sizes. The batch size was set to 16; furthermore, the GridsearchCV method of the Sklearn library was used to find the best set of hyperparameters. Finally, Adam was used as the optimization algorithm that was suggested by Kingma and Ba (2014). As a final step, the slots of the utterances generated by the seq2seq model were filled with the vocabulary created in the sentence delexicalization method. Using the seq2seq data augmentation model, a total of 8,320 synthetic sentences were generated.

These 8,320 synthetic sentences were used to improve the language model developed by Cardenas et al. (2018). The language model was developed by taking the modified Kneser-Ney n-gram model (Heafield et al., 2013). In addition, given the high presence of morphologically rare words, a singleton pruning rate with a "κ" of 0.04 (Cárdenas et al, 2018) was used to randomly replace only a "κ" fraction of the once-occurring words in the training data with a global UNK symbol. We succeeded in developing a word-level 4-gram language model with a perplexity of 282.45.

---

[14] https://github.com/rjzevallos/Data-Augmentation-for-Quechua

As a second step of our new data augmentation method, a Tacotrom2-based[15] TTS model for Quechua was developed (Shen et al., 2018). This model was implemented using the 99 hours of audio from the Southern Quechua training dataset.

The model was trained using the steps from the Spanish Tacotrom2 repository[16]. The model was trained in 1500 epochs with 1 GPU, the learning rate was 0.001, a batch size of 32 was used, the Adam optimizer was also used and finally we applied a gradient clipping threshold of 0.1. The ASR model architecture consists of 1 embedding layer, 3 convolutional layers, 1 Bi-LSTM layer, 1 LSA layer, 2 pre-net layers and 5 post-net convolutional layers. Finally, the convolutional layers have a dropout of 0.5 and the Bi-LSTM layers have a zoneout of 0.1. The Quechua TTS model achieved a mean opinion score (MOS) of 3.15% which was obtained by consulting Quechua speakers from southern Peru.

Finally, the 8,320 synthetic sentences generated in the first step of our method were used to generate synthetic speech using the TTS model developed earlier. In total, 99 hours of synthetic audio were generated.

Both the 8,320 synthetic sentences with the 99 hours of synthetic audio were aligned to train the ASR model. The ASR model was trained with 99 hours of original transcribed audio and 99 hours of synthetic transcribed audio. To train the ASR model, the steps of the first experiment were followed in addition to using the new language model developed in this experiment. The ASR model trained in this experiment achieved a WER of 22.75%.

---

[15] https://github.com/mozilla/TTS
[16] https://github.com/mozilla/TTS/wiki/TTS-Notebooks-and-Tutorials

### 4.3.4. Fourth experiment

In the fourth experiment, the ASR model was trained following the same steps of the first experiment but doubling the 99 hours of transcribed audio i.e. 198 hours of original transcribed audio. In addition, the language model of Cardenas et al. (2018) was used. The ASR model obtained a WER of 26.14%.

## 5. Results

In this investigation, fourth experiments were conducted to evaluate the new data augmentation method using a seq2seq model to generate synthetic text and a TTS model to generate synthetic speech, in order to improve the result of the ASR model of Southern Quechua. The results of the experiments are shown in the following table.

**Table 4** Comparison of the results of the data augmentation

| Experiment | Training data | Training hours | WER (%) |
|---|---|---|---|
| Exp 1 | natural (Baseline) | 99 | 31.48 |
| Exp 2 | natural + distorted audio | 99 + 99 | 25.13 |

| Exp 3 | natural + synthetic audio and synthetic text aligned + new LM | 99 + 99 | **22.75** |
| Exp 4 | natural + natural | 99 + 99 | 26.14 |

The first experiment was performed as a baseline to compare the data augmentation methods performed in the following experiments. The baseline ASR model used 99 hours of transcribed audio achieving a WER of 31.48%.

In the second experiment, the data augmentation method based on velocity distortion was used. The ASR model trained with 99 hours of original transcribed audio and 99 hours of distorted transcribed audio managed to decrease the WER from 31.48% to 25.13%. Therefore, a relative improvement of 6.35% in WER was obtained by augmenting data based on velocity distortion in the transcribed audios.

In the third experiment, we evaluated the new data augmentation system using the seq2seq model and the TTS model, in order to create synthetic texts to improve the language model and at the same time, these synthetic texts served as input to generate synthetic speech in the TTS model. In this experiment, the WER of the ASR model decreased from 31.48% to 22.75%. Therefore, a relative improvement of 8.73% in WER was obtained.

In the fourth experiment, the original data were duplicated without any modification, just to verify whether natural data augmentation is more efficient than previous data augmentation methods. In this experiment, the WER of the ASR model was reduced from 31.48% to 26.14%. Therefore, a relative improvement of 5.34% in WER was obtained.

## 5.1. Ablation

In order to better understand each step of our data augmentation method, we performed an ablation test. Each of the two parts of our method is removed respectively, including synthetic text generation using the seq2seq model and synthetic audio generation using the TTS model.

The results of the ASR model using each step of our data augmentation method are shown below.

**Table 5** Results of the ASR model by removing each step of our DA method

| Method | Training data | LM | Training hours | WER (%) |
|---|---|---|---|---|
| Our data augmentation method | natural + synthetic audio and synthetic text aligned | New LM | 99 + 99 | 22.75 |
| Without TTS model | Natural | New LM | 99 | 29.71 |
| Without Seq2seq model | natural + synthetic audio | Old LM | 99 + 99 | 25.64 |

In the case of our data augmentation method without using the TTS model to generate synthetic audio, a 6.96% increase in WER is observed compared to the result of our full data augmentation method. Since only the synthetic text generation method was used to improve the language model, this ablation shows that our data augmentation without using the TTS model to generate synthetic audio impacts significantly in improving the ASR model.

For our data augmentation method without using seq2seq model to generate synthetic text, a 2.89% increase in WER is observed compared to the result of our full data augmentation method. Given that only the original text with the TTS model was used to generate synthetic audio, we attribute that the little increase in WER in the results is that the seq2seq model generating synthetic text to improve the language model has no major effect on the improvement of the ASR model.

Finally, we can say that the step that has the most effect within our new data augmentation system is the TTS model to generate synthetic audio. However, it is not known whether using synthetic text to generate synthetic audio using the TTS model and not using the enhanced language model has as much effect as the TTS model to generate synthetic audio from original text.

# 6. Result analysis

The new data augmentation method was compared with other data augmentation methods to verify if it could obtain better results for the ASR model of Quechua.

The most notable result was the third experiment, in which our new data augmentation method based on a seq2seq model was used to generate synthetic texts and a TTS model was used to generate synthetic speech.

The figure below shows the effect of this new data augmentation method on the corpus.

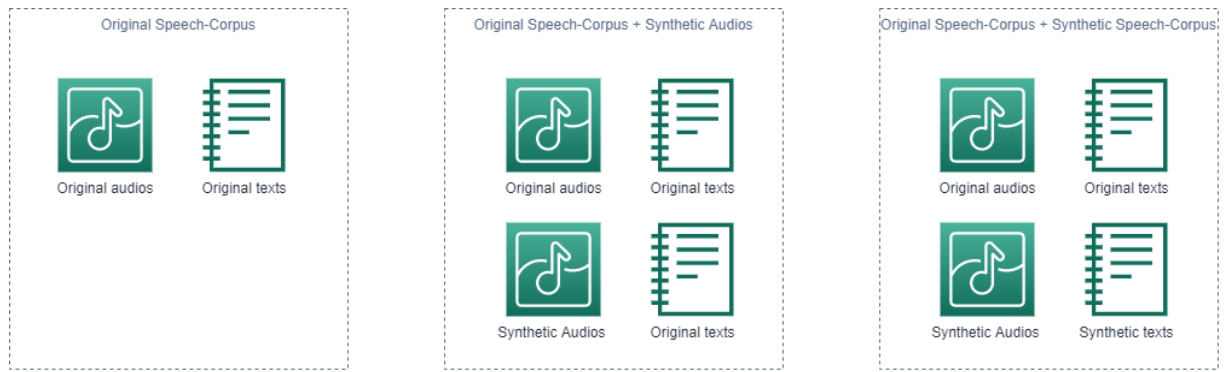

**Figure 6** Variation of the speech-corpus according to the data augmentation technique.

To simplify the analysis, the results of the fifth experiment can be divided into the following parts:

- The seq2seq model generated synthetic texts for the binding languages efficiently.
- The synthetic texts obtained by the seq2seq model improved the language model.
- The language model trained with the synthetic texts improved the performance of the ASR model.
- The synthetic audios generated by the TTS model from the synthetic texts improved the performance of the ASR model.
- A new corpus of transcribed speech with original and synthetic data was obtained.

Finally, it can be seen that our data augmentation system generates more efficient synthetic text and audio for the ASR model of Quechua, since in the fourth experiment, it could be observed that by duplicating the original data to train the ASR model, the improvement is lower than training the model with synthetic data, demonstrating that our data augmentation method for agglutinative languages is more efficient.

# 7. Conclusion

In this research, a data augmentation system was developed to improve the results of ASR models for agglutinative and low-resource languages such as Quechua. A delexicalization method for agglutinative languages was built as part of the data augmentation seq2seq model; furthermore, a language model was developed with the new synthetic texts generated by the data augmentation seq2seq model; also, a TTS based on the Tacotron 2 model was built to generate synthetic audio and finally an ASR based on the Wav2letter model was developed.

Some experiments were performed with the original training data and with the augmented data using the different methods in order to compare the performance of the new data augmentation method.

The results showed that the data augmentation technique works well for improving speech recognition systems in the case of agglutinative and low-resource languages. In this case, a relative improvement of 8.73% of WER was obtained using the combination of text data augmentation and synthetic speech data augmentation.

# 8. References

Aimituma Suyo, F., & Churata Urtado, R. M. (2019). Conversor de voz a texto para el idioma quechua usando la herramienta de reconocimiento de voz KALDI y una red neuronal profunda.

Amodei, D., Anubhai, R., Battenberg, E., Case, C., Casper, J., Catanzaro, B., ... & Zhu, Z. (2016). End to end speech recognition in English and Mandarin.

Andrew L. Maas, Awni Y. Hannun, Daniel Jurafsky, and Andrew Y. Ng, (2014) .First-pass large vocabulary continuous speech recognition using bi-directional recurrent dnns, CoRR abs/1408.2873.

Austin, P. K., & Sallabank, J. (Eds.). (2011). The Cambridge handbook of endangered languages. Cambridge University Press.

Bird, S. (2020, December). Decolonising Speech and Language Technology. In Proceedings of the 28th International Conference on Computational Linguistics (pp. 3504-3519).

Black, A. W., & Taylor, P. A. (1997). Automatically clustering similar units for unit selection in speech synthesis.

Camacho Caballero, L., & Zevallos Salazar, R. (2020). Lingüística computacional para la revitalización y el poliglotismo. Letras (Lima), 91(134), 184-198.

Cardenas, R., Zevallos, R., Baquerizo, R., & Camacho, L. (2018). Siminchik: A speech corpus for preservation of southern quechua. ISI-NLP 2, 21.

Cerrón-Palomino, Rodolfo. (1994). Quechua sureño, diccionario unificado quechua-castellano, castellano-quechua. Biblioteca Nacional del Perú, Lima.

Chacca Chuctaya, H. F. (2019). Isolated Automatic Speech Recognition of Quechua Numbers using MFCC, DTW AND KNN Reconocimiento Automático de Habla Aislado de Números en Quechua usando MFCC, DTW AND KNN.

Chen, Stanley F., Brian Kingsbury, Lidia Mangu, Daniel Povey, George Saon, Hagen Soltau and Geoffrey Zweig. (2006). Advances in speech transcription at IBM under the DARPA EARS program. IEEE Trans. Audio, Speech & Language Processing 14: 1596-1608.

Chen, Y. N., Hakanni-Tür, D., Tur, G., Celikyilmaz, A., Guo, J., & Deng, L. (2016, December). Syntax or semantics? Knowledge-guided joint semantic frame parsing. In 2016 IEEE Spoken Language Technology Workshop (SLT) (pp. 348-355). IEEE.

Collobert, R., Puhrsch, C., & Synnaeve, G. (2016). Wav2letter: an end-to-end convnet-based speech recognition system. arXiv preprint arXiv:1609.03193.

Cui, X., Goel, V., & Kingsbury, B. (2015). Data augmentation for deep neural network acoustic modeling. IEEE/ACM Transactions on Audio, Speech, and Language Processing, 23(9), 1469-1477.

Fader, A., Zettlemoyer, L., & Etzioni, O. (2013, August). Paraphrase-driven learning for open question answering. In Proceedings of the 51st Annual Meeting of the Association for Computational Linguistics (Volume 1: Long Papers) (pp. 1608-1618).

Godfrey, J. J., Holliman, E. C., & McDaniel, J. (1992, March). SWITCHBOARD: Telephone speech corpus for research and development. In Acoustics, Speech, and Signal

Processing, IEEE International Conference on (Vol. 1, pp. 517-520). IEEE Computer Society.

Gokay, R., & Yalcin, H. (2019, March). Improving low Resource Turkish speech recognition with Data Augmentation and TTS. In 2019 16th International Multi-Conference on Systems, Signals & Devices (SSD) (pp. 357-360). IEEE.

Graves, Alex, Santiago Fernández, Faustino J. Gomez and Jürgen Schmidhuber. (2006). Connectionist temporal classification: labelling unsegmented sequence data with recurrent neural networks. ICML.

Gulati, A., Qin, J., Chiu, C. C., Parmar, N., Zhang, Y., Yu, J., ... & Pang, R. (2020). Conformer: Convolution-augmented transformer for speech recognition. arXiv preprint arXiv:2005.08100.

Gulati, A., Qin, J., Chiu, C. C., Parmar, N., Zhang, Y., Yu, J., ... & Pang, R. (2020). Conformer: Convolution-augmented transformer for speech recognition. arXiv preprint arXiv:2005.08100.

Hannun, A. Y., Maas, A. L., Jurafsky, D., & Ng, A. Y. (2014). First-pass large vocabulary continuous speech recognition using bi-directional recurrent DNNs. arXiv preprint arXiv:1408.2873.

Hannun, A., Case, C., Casper, J., Catanzaro, B., Diamos, G., Elsen, E., & Ng, A. Y. (2014). Deep speech: Scaling up end-to-end speech recognition. arXiv preprint arXiv:1412.5567.

Heafield, K., Pouzyrevsky, I., Clark, J. H., & Koehn, P. (2013, August). Scalable modified Kneser-Ney language model estimation. In Proceedings of the 51st Annual

Meeting of the Association for Computational Linguistics (Volume 2: Short Papers) (pp. 690-696).

Heggarty, P., Valko, M. L., Huarcaya, S. M., Jerez, O., Pilares, G., Paz, E. P., ... & Usandizaga, H. (2005). Enigmas en el origen de las lenguas andinas: aplicando nuevas técnicas a las incógnitas por resolver. Revista Andina, 40, 9-57.

Hinton, G. E., Krizhevsky, A., & Sutskever, I. (2012). Imagenet classification with deep convolutional neural networks. Advances in neural information processing systems, 25, 1106-1114.

Hou, Y., Liu, Y., Che, W., & Liu, T. (2018). Sequence-to-sequence data augmentation for dialogue language understanding. arXiv preprint arXiv:1807.01554.

Hsu, J. Y., Chen, Y. J., & Lee, H. Y. (2020, May). Meta learning for end-to-end low-resource speech recognition. In ICASSP 2020-2020 IEEE International Conference on Acoustics, Speech and Signal Processing (ICASSP) (pp. 7844-7848). IEEE.

Hunt, A. J., & Black, A. W. (1996, May). Unit selection in a concatenative speech synthesis system using a large speech database. In 1996 IEEE International Conference on Acoustics, Speech, and Signal Processing Conference Proceedings (Vol. 1, pp. 373-376). IEEE.

Kallasjoki, H., Remes, U., Gemmeke, J. F., Virtanen, T., & Palomäki, K. J. (2011). Uncertainty measures for improving exemplar-based source separation. In Twelfth Annual Conference of the International Speech Communication Association.

Kim, Y., Jernite, Y., Sontag, D., & Rush, A. (2016, March). Character-aware neural language models. In Proceedings of the AAAI conference on artificial intelligence (Vol. 30, No. 1).

Kingma, D. P., & Ba, J. (2014). Adam: A method for stochastic optimization. arXiv preprint arXiv:1412.6980.

Klein, G., Kim, Y., Deng, Y., Senellart, J., & Rush, A. M. (2017). Opennmt: Open-source toolkit for neural machine translation. arXiv preprint arXiv:1701.02810.

Kobayashi, S. (2018). Contextual augmentation: Data augmentation by words with paradigmatic relations. arXiv preprint arXiv:1805.06201.

Łańcucki, A. (2021, June). Fastpitch: Parallel text-to-speech with pitch prediction. In ICASSP 2021-2021 IEEE International Conference on Acoustics, Speech and Signal Processing (ICASSP) (pp. 6588-6592). IEEE.

Landerman, P. N. (1991). Quechua dialects and their classification (Doctoral dissertation, University of California, Los Angeles).

Lu, Q., Li, Y., Qin, Z., Liu, X., & Xie, Y. (2020, May). Speech recognition using efficientnet. In Proceedings of the 2020 5th International Conference on Multimedia Systems and Signal Processing (pp. 64-68).

Luong, M. T., Pham, H., & Manning, C. D. (2015). Effective approaches to attention-based neural machine translation. arXiv preprint arXiv:1508.04025.

Mager, M., Gutierrez-Vasques, X., Sierra, G., & Meza, I. (2018). Challenges of language technologies for the indigenous languages of the Americas. arXiv preprint arXiv:1806.04291.

Maguiño-Valencia, D., Oncevay, A., & Cabezudo, M. A. S. (2018, May). WordNet-SHP: Towards the building of a lexical database for a Peruvian minority language. In Proceedings of the Eleventh International Conference on Language Resources and Evaluation (LREC 2018).

Maguiño-Valencia, D., Oncevay, A., & Cabezudo, M. A. S. (2018, May). WordNet-SHP: Towards the building of a lexical database for a Peruvian minority language. In Proceedings of the Eleventh International Conference on Language Resources and Evaluation (LREC 2018).

Monson, C., Llitjós, A. F., Aranovich, R., Levin, L., Brown, R., Peterson, E., & Lavie, A. (2006). Building NLP systems for two resource-scarce indigenous languages: Mapudungun and Quechua. Strategies for developing machine translation for minority languages, 15.

Morin, F., & Bengio, Y. (2005, January). Hierarchical probabilistic neural network language model. In Aistats (Vol. 5, pp. 246-252).

Nguyen, T. S., Stueker, S., Niehues, J., & Waibel, A. (2020, May). Improving sequence-to-sequence speech recognition training with on-the-fly data augmentation. In ICASSP 2020-2020 IEEE International Conference on Acoustics, Speech and Signal Processing (ICASSP) (pp. 7689-7693). IEEE.

Park, D. S., Chan, W., Zhang, Y., Chiu, C. C., Zoph, B., Cubuk, E. D., & Le, Q. V. (2019). Specaugment: A simple data augmentation method for automatic speech recognition. arXiv preprint arXiv:1904.08779.

Paul, D. B., & Baker, J. (1992). The design for the Wall Street Journal-based CSR corpus. In Speech and Natural Language: Proceedings of a Workshop Held at Harriman, New York, February 23-26, 1992.

Pratap, V., Hannun, A., Xu, Q., Cai, J., Kahn, J., Synnaeve, G., ... & Collobert, R. (2019, May). Wav2letter++: A fast open-source speech recognition system. In ICASSP 2019-2019 IEEE International Conference on Acoustics, Speech and Signal Processing (ICASSP) (pp. 6460-6464). IEEE.

Panayotov, V., Chen, G., Povey, D., & Khudanpur, S. (2015, April). Librispeech: an asr corpus based on public domain audio books. In 2015 IEEE international conference on acoustics, speech and signal processing (ICASSP) (pp. 5206-5210). IEEE.

Ragni, A., Knill, K. M., Rath, S. P., & Gales, M. J. (2014, September). Data augmentation for low resource languages. In INTERSPEECH 2014: 15th Annual Conference of the International Speech Communication Association (pp. 810-814). International Speech Communication Association (ISCA).

Rios, A. (2015). A basic language technology toolkit for quechua (Doctoral dissertation, University of Zurich).

Rios, A. (2015). A basic language technology toolkit for Quechua (Doctoral dissertation, University of Zurich).

Rios, A., & Castro Mamani, R. (2014). Morphological disambiguation and text normalization for southern quechua varieties.

Rios, A., Göhring, A., & Volk, M. (2008). A Quechua-Spanish parallel treebank. LOT Occasional Series, 12, 53-64.

Rossenbach, N., Zeyer, A., Schlüter, R., & Ney, H. (2020, May). Generating synthetic audio data for attention-based speech recognition systems. In ICASSP 2020-2020 IEEE International Conference on Acoustics, Speech and Signal Processing (ICASSP) (pp. 7069-7073). IEEE.

Rudnick, A. (2011, September). Towards Cross-Language Word Sense Disambiguation for Quechua. In Proceedings of the Second Student Research Workshop associated with RANLP 2011 (pp. 133-138).

Rygaard, L. V. (2015). Using Synthesized Speech to Improve Speech Recognition for Low–Resource Languages. Retrieved October, 8, 2018.

S. Matsoukas, J. -L. Gauvain, G. Adda, T. Colthurst, Chia-Lin Kao, O. Kimball, L. Lamel, F. Lefevre, J. Z. Ma, J. Makhoul, L. Nguyen, R. Prasad, R. Schwartz, H. Schwenk, and Bing Xiang. (2006). Advances in transcription of broadcast news and conversational telephone speech within the combined EARS BBN/LIMSI system. Trans. Audio, Speech and Lang. Proc. 14, 5.

Seide, F., Li, G., & Yu, D. (2011). Conversational speech transcription using context-dependent deep neural networks. In Twelfth annual conference of the international speech communication association.

Shen, J., Pang, R., Weiss, R. J., Schuster, M., Jaitly, N., Yang, Z., ... & Wu, Y. (2018, April). Natural tts synthesis by conditioning wavenet on mel spectrogram predictions. In 2018 IEEE International Conference on Acoustics, Speech and Signal Processing (ICASSP) (pp. 4779-4783). IEEE.

Shen, J., Pang, R., Weiss, R. J., Schuster, M., Jaitly, N., Yang, Z., ... & Wu, Y. (2018, April). Natural tts synthesis by conditioning wavenet on mel spectrogram predictions.
45

In 2018 IEEE International Conference on Acoustics, Speech and Signal Processing (ICASSP) (pp. 4779-4783). IEEE.

Sotelo, J., Mehri, S., Kumar, K., Santos, J. F., Kastner, K., Courville, A., & Bengio, Y. (2017). Char2wav: End-to-end speech synthesis.

Sutskever, I., Vinyals, O., & Le, Q. V. (2014). Sequence to sequence learning with neural networks. Advances in neural information processing systems, 27, 3104-3112.

Synnaeve, G., Xu, Q., Kahn, J., Likhomanenko, T., Grave, E., Pratap, V., ... & Collobert, R. (2019). End-to-end asr: from supervised to semi-supervised learning with modern architectures. arXiv preprint arXiv:1911.08460.

Tokuda, K., Yoshimura, T., Masuko, T., Kobayashi, T., & Kitamura, T. (2000, June). Speech parameter generation algorithms for HMM-based speech synthesis. In 2000 IEEE International Conference on Acoustics, Speech, and Signal Processing. Proceedings (Cat. No. 00CH37100) (Vol. 3, pp. 1315-1318). IEEE.

Torero, A. (1964). Los dialectos quechua

Vargas, J., Cruz, J., & Castro, R. (2012). Let's Speak Quechua: The Implementation of a Text-to-Speech System for the Incas' Language, IberSPEECH 2012 – VII Jornadas en Tecnología del Habla and III Iberian SLTech Workshop.

Wang, Y., Skerry-Ryan, R. J., Stanton, D., Wu, Y., Weiss, R. J., Jaitly, N., ... & Saurous, R. A. (2017). Tacotron: Towards end-to-end speech synthesis. arXiv preprint arXiv:1703.10135.

Wei, J., & Zou, K. (2019). Eda: Easy data augmentation techniques for boosting performance on text classification tasks. arXiv preprint arXiv:1901.11196.

Xie, B., Passonneau, R., Wu, L., & Creamer, G. G. (2013, August). Semantic frames to predict stock price movement. In Proceedings of the 51st annual meeting of the association for computational linguistics (pp. 873-883).

Yoshimura, T., Hayashi, T., Takeda, K., & Watanabe, S. (2020, May). End-to-end automatic speech recognition integrated with ctc-based voice activity detection. In ICASSP 2020-2020 IEEE International Conference on Acoustics, Speech and Signal Processing (ICASSP) (pp. 6999-7003). IEEE.

Yu, A. W., Dohan, D., Luong, M. T., Zhao, R., Chen, K., Norouzi, M., & Le, Q. V. (2018). Qanet: Combining local convolution with global self-attention for reading comprehension. arXiv preprint arXiv:1804.09541.

Zaremba, W., Sutskever, I., & Vinyals, O. (2014). Recurrent neural network regularization. arXiv preprint arXiv:1409.2329.

Zen, H., Tokuda, K., & Black, A. W. (2009). Statistical parametric speech synthesis. Speech communication, 51(11), 1039-1064.

Zevallos, R., Cordova, J., & Camacho, L. (2019, August). Automatic Speech Recognition of Quechua Language Using HMM Toolkit. In Annual International Symposium on Information Management and Big Data (pp. 61-68). Springer, Cham.

Zhang, X., Zhao, J., & LeCun, Y. (2015). Character-level convolutional networks for text classification. Advances in neural information processing systems, 28, 649-657.